\documentclass[conference]{IEEEtran}
\IEEEoverridecommandlockouts
\usepackage{cite}
\usepackage{amsmath,amssymb,amsfonts}
\usepackage{graphicx}
\usepackage{textcomp}
\usepackage{amsmath}
\usepackage{algorithm}
\usepackage[noend]{algpseudocode}
\usepackage{caption}
\usepackage{subcaption}
\usepackage{array}
\usepackage[export]{adjustbox}
\usepackage{soul}
\usepackage{helvet}
\usepackage{booktabs}

\usepackage{acro}
\DeclareRobustCommand{\acrodef}[2]{\DeclareAcronym{#1}{short=#1,long=#2}}
\acrodef{IQ}{in-phase and quadrature}
\acrodef{PDUs}{Protocol Data Units}
\acrodef{RRC}{Radio Resource Control}
\acrodef{ZMQ}{ZeroMQ}
\acrodef{LSTM}{Long Short-Term Memory}
\acrodef{LAL}{Listen-and-Learn}
\acrodef{TS}{Technical Specifications}
\acrodef{NR}{New Radio}
\acrodef{eMBB}{enhancing Mobile Broadband}
\acrodef{mMTC}{massive Machine Type Communications}
\acrodef{O-RAN}{Open Radio Access Network}
\acrodef{NAS}{Non-Access Stratum}
\acrodef{PHY}{physical}
\acrodef{AS}{Access Stratum}
\acrodef{UE}{User Equipment}
\acrodef{MITM}{man-in-the-middle}
\acrodef{OTA}{over-the-air}
\acrodef{MAC}{medium access control}
\acrodef{CN}{core network}
\acrodef{gNB}{gNodeB}
\acrodef{DoS}{Deny of Service}
\acrodef{BS}{Base Station}
\acrodef{NSA}{Non Standard-Alone}
\acrodef{SA}{Standard-Alone}
\acrodef{EPS}{Evolved Packet System}
\acrodef{eNB}{evolved NodeBs}
\acrodef{AKA}{authentication and key agreement}
\acrodef{EC-AKA}{Ensured confidentiality Authentication and Key agreement}
\acrodef{ASME}{Access Security Management Entity}
\acrodef{UP}{User Plane}
\acrodef{KDF}{key derivation function}
\acrodef{IMSI}{international mobile subscriber identity}
\acrodef{S-TMSI}{SAE Temporary Mobile Subscriber Identity}
\acrodef{OAI}{Open Air Interface}
\acrodef{NFV}{Network Function Virtualization}
\acrodef{SDR}{Software Defined Radios}
\acrodef{SUT}{System Under Test}

\usepackage{xcolor}
\def\BibTeX{{\rm B\kern-.05em{\sc i\kern-.025em b}\kern-.08em
    T\kern-.1667em\lower.7ex\hbox{E}\kern-.125emX}}
\begin{document}

\title{From Ambiguity to Explicitness:  NLP-Assisted 5G Specification Abstraction for Formal Analysis}

\author{\IEEEauthorblockN{Shiyu Yuan\IEEEauthorrefmark{1}, Jingda Yang\IEEEauthorrefmark{1}, 
Sudhanshu Arya\IEEEauthorrefmark{1},
Carlo Lipizzi\IEEEauthorrefmark{1}  and Ying Wang\IEEEauthorrefmark{1}}\\
\IEEEauthorblockA{\IEEEauthorrefmark{1}Stevens Institute of Technology, Hoboken, NJ \\
Email: \IEEEauthorrefmark{1}\{syuan14, jyang76, sarya, clipizzi, ywang6\}@stevens.edu}}

\maketitle

\begin{abstract}
Formal method-based analysis of the 5G Wireless Communication Protocol is crucial for identifying logical vulnerabilities and facilitating an all-encompassing security assessment, especially in the design phase. Natural Language Processing (NLP) assisted techniques and most of the tools are not widely adopted by the industry and research community. Traditional formal verification through a mathematics approach heavily relied on manual logical abstraction prone to being time-consuming, and error-prone. The reason that the NLP-assisted method did not apply in industrial research may be due to the ambiguity in the natural language of the protocol designs nature is controversial to the explicitness of formal verification. To address the challenge of adopting the formal methods in protocol designs, targeting (3GPP) protocols that are written in natural language, in this study, we propose a hybrid approach to streamline the analysis of protocols. We introduce a two-step pipeline that first uses NLP tools to construct data and then uses constructed data to extract identifiers and formal properties by using the NLP model. The identifiers and formal properties are further used for formal analysis. We implemented three models that take different dependencies between identifiers and formal properties as criteria. Our results of the optimal model reach valid accuracy of 39\% for identifier extraction and 42\% for formal properties predictions. Our work is proof of concept for an efficient procedure in performing formal analysis for large-scale complicate specification and protocol analysis, especially for 5G and nextG communications. 


\end{abstract}

\begin{IEEEkeywords}
5G, formal dependency table, NLP, identifier, formal property, language model
\end{IEEEkeywords}


\section{Introduction}
The fifth-generation (5G) technology introduces more flexibility than previous generations of cellular networks for novel functionalities and higher performance. It also enables novel customization when applied to various verticals. Meanwhile, the flexibility and customization, the design of more complicated specifications and protocols, along with amendment agreements bring forth the need for more sophisticated protocol examinations and verification to detect functional and nonfunctional vulnerabilities inherited from previous generations of networks and newly introduced. To ensure the security of these specifications and protocols, fuzz testing, and formal verification are employed to detect vulnerabilities in the system and verify the accuracy of implemented protocol behaviors.

To expedite the assessment of protocol security, formal verification is employed to mathematically model identifiers and properties, using formal methods and fuzz testing to reason about and establish their security \cite{yang2023formal,yang2023formal_guided,basin2018formal}.Fuzz testing is an effective method that utilizes a plethora of random or mutated inputs to uncover system vulnerabilities. This automated approach significantly reduces labor costs associated with manual testing efforts. However, its effectiveness relies on the quality and diversity of test inputs generated \cite{yang2023formal, yang20235g}. 

In the realm of 5G Wireless Communication Protocol, formal verification plays a pivotal role in ensuring the reliability, security, interoperability, and performance of 5G networks, including protocol correctness, security measures, and overall system robustness~\cite{hussain20195greasoner}. Unlike automated physical simulation testing approaches, formal verification is a labor-intensive process that necessitates manual expertise in extracting protocols' procedures and relationships. Consequently, there is a pressing need for an automated system capable of analyzing protocols, extracting identifiers, and establishing their relationships.


Recently, an NLP based 5G vulnerability detection approach has been proposed \cite{wang2023nlp}. The cross-layer 5G vulnerability detection is conducted via fuzzing-generated run-time profiling with piloting on srsRAN. To characterize the performance, the authors leverage various machine learning classification algorithms, including K-Nearest Neighbors, Random Forest, and Logistic Regression with performance reaching as high as $\approx 95.9\%$. Although the approach offers high accuracy, it relies on generating random or semi-random inputs to the target system, thereby may not be able to explore all possible code paths. This can result in missing certain vulnerabilities that require specific sequences of actions to trigger. Moreover, due to the complex 5G network, fuzzing can lead to a massive number of test cases, making it challenging to handle the state-space explosion problem effectively.

\section{Related Work}
\noindent\textbf{NextG Wireless Communication Protocol and Formal Verification:}
With the rapid development of wireless communication technologies and the ongoing efforts to address vulnerabilities in previously defined protocols, experts regularly introduce new protocols that modify high-risk rules and introduce additional constraints. However, the comprehension of these newly released protocols by human experts remains a significant challenge in efficiently transferring them into formal properties for cybersecurity purposes. Therefore, there is a growing need for an automatic framework that can extract identifiers and analyze formal relationships, ensuring the security of protocols while alleviating the burden of manual analysis. 
Formal verification has emerged as a valuable tool in numerous research studies for detecting logical vulnerabilities and ensuring error-free systems with propriety \cite{hussain20195greasoner}. In contrast to traditional fuzz testing, formal verification leverages mathematical proofs to rigorously verify the logical correctness of procedures. By focusing on the properties and relationships of each identifier, formal verification offers a systematic approach to ensure the integrity and reliability of system behavior. However, due to the stringent nature of mathematical proofs, formal verification typically requires the expertise of security professionals for manual analysis. The extensive protocol analysis and recurring context updates necessitate the development of efficient protocol analyzer tools to streamline the extraction of identifiers and their properties in the field of cybersecurity \cite{zhang2019formal}.

\noindent\textbf{Identifier and Formal Property Labeling:}
Identifiers in communication protocols often represent specific protocol components, each with a unique role in ensuring efficient and effective communication. The network identifiers in 5G system are divided into subscriber identifiers and UE identifiers \cite{3gpp.38.331}. These identifiers could be protocol names (e.g., `5G', `LTE', `TCP'), specific algorithm names (e.g., `RSA', 'AES'), or other technical terminologies unique to the communication domain (e.g., `RCCRequest', `RCCConfiguration'). Simultaneously, formal properties associated with these identifiers refer to the mathematically definable characteristics that describe the behavior of the protocols. These might include attributes like `Authentication', `Accounting',`Integrity',`Confidentiality' \cite{yang2023formal}. Together, the identifier and its formal properties provide a detailed specification for a protocol component, outlining its function, behavior, and performance within the communication system. The extraction of formal verification information provides a foundational step towards a deeper understanding and analysis of complex communication protocols such as 5G. 

\noindent\textbf{Information Extraction in Natural Language Processing:}
Information Extraction (IE) is a task in NLP that involves automatically extracting structured information from unstructured text data \cite{jurafskyspeech}. This typically involves processing text and identifying relevant pieces of information, such as entities, relationships between entities, and specific attributes of entities or relationships. IE has three main tasks, named entity recognition (NER), semantic role labeling (SRL), and relation extraction (RE) \cite{khetan2021knowledge} with two main approaches on these three tasks, a heuristic approach based on the transition-based model and supervised learning through a neural network and annotation data \cite{yuan2023information}. Despite the availability of generic-domain datasets for IE, domain-specific IE resources, such as annotated datasets for NER or RE, are scarce. This scarcity presents a significant challenge when executing IE tasks within a specific domain. Several studies provide NER pipeline such as in work \cite{alam2022cyner}, but their training corpus are more general cybersecurity content instead of communication protocols. 

\subsubsection{NER}
NER is to identify named entities or semantic roles in a sequence of tokens \cite{jurafskyspeech}. In this study, our focus is on domain-specific terminology in 5G communication protocol, also referred to as 'identifiers'. The focus of this study is to undertake a systematic analysis of protocol documents, with the specific aim of locating and extracting the pertinent identifiers contained within them and predicting the formal properties that delineate the relationships between the extracted identifiers.
\subsubsection{RE}
RE is a task in information extraction that involves identifying and classifying semantic relationships between entities in text. These entities are typically pre-identified, and the goal of relation extraction is to ascertain how these entities are connected \cite{banko2008tradeoffs}. However, in our task, the entities are not pre-identified. Instead, the entities need to be located and extracted during the training phase, which adds a layer of complexity to the task. In the context of our study, relations are considered as formal properties between identifiers.

To address the research gap, streamline the analysis pipeline, and mitigate the potential risks of experimental errors, we propose a unique solution leveraging natural language processing techniques to analyze the 5G protocol. The protocol, primarily written in natural language, serves as the basis for our solution. Our approach specifically integrates two different natural language processing methodologies for the analysis. Initially, we employ a traditional heuristic technique utilizing information extraction methods to identify named entities and semantic roles. Subsequently, we implement a pre-trained language model to extract the entities or identifiers and the formal property between the identifiers within these sentences. 

\noindent\textbf{Contributions: }
To this end, we make the following contributions:
\begin{itemize}
    \item This work presents a first-of-its-kind and a new approach to doing formal analysis employing the NLP technique. The proposed approach aims to overcome the limitations imposed by the use of the conventional methods, thereby allowing wider state-space explosion and coverage. We elicit the set of properties from the informal standardization documents and encode them as analyzable queries in the formal model.
    \item The presented method eliminates the need for resource-intensive physical experiments by utilizing NLP techniques. This approach promotes an automated and efficient protocol analysis pipeline, reducing reliance on specialized setup and equipment while minimizing the risk of error. Consequently, our method significantly improves the accessibility and scalability of the analysis.
    \item In particular, a novel approach to doing formal analysis employing two complementary NLP solutions is presented. The first, a traditional heuristic, uses information extraction techniques to identify named entities (identifiers) and semantic roles within the protocol text. The second leverages a pre-trained language model to extract corresponding entities (identifiers) and formal properties within the sentences. 
    
\end{itemize}


In summary, our proposed method offers a novel, automated, and efficient approach to 5G Wireless Communication Protocol analysis. By leveraging NLP techniques, we can analyze the protocol more efficiently, and intuitively, addressing several of the limitations inherent in current approaches.\\

The remainder of this paper is structured as follows. Section 3 provides a comprehensive overview of the methodologies and the algorithms we developed for this study. The experimental setup and the evaluation metrics used for the study are detailed in Section 4. The results and analysis of these experiments are presented in Section 5. We provide a discussion and interpretation of these results in Section 6. Finally, the conclusions are drawn in Section 7.

\section{Methodology}
Traditional IE tasks typically rely on annotated datasets for NER, with entities for RE tasks being pre-defined. However, our research encounters two major obstacles: the scarcity of datasets and the unavailability of pre-defined entities for RE tasks. To address these challenges, we propose a novel method.

\noindent\textbf{(1)} Regarding the annotated dataset, domain experts have noted that tables within these protocols are rich in domain-specific terminologies, most of which are the identifiers we aim to extract. We have leveraged the inherent structure of these tables embedded in the protocols as a natural means to extract these identifiers.

\noindent\textbf{(2)} Addressing the issue of unavailable pre-defined entities in the RE task, we proposed a pipeline capable of concurrently extracting identifiers and relations (also considered as formal properties) during the training process.

\begin{figure*}[h!]
\centering
\includegraphics[width=5.3in]{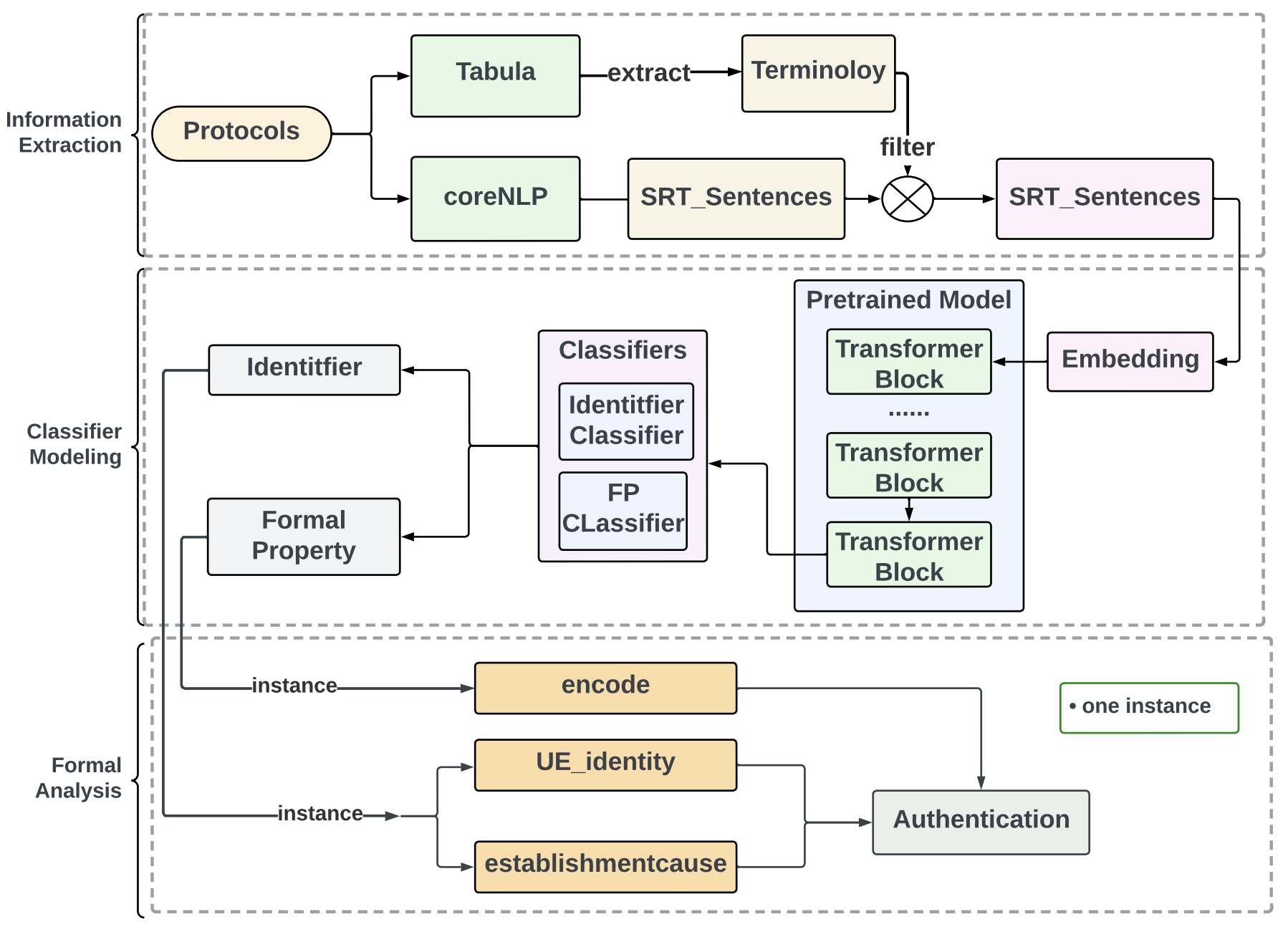}
\caption{System Diagram}
\label{fig:systemidagram}
\vspace{-2mm}
\end{figure*}

The whole system is illustrated in Figure. \ref{fig:systemidagram} which includes two components of Information Extraction and Formal Analysis . 

Based on the constructed dataset, we proposed a hybrid joint training model based on pretrained language model. Specifically, we first locate and extract the identifiers and use the extracted identifiers and the context sentence as latent representation to classify the relations between the identifiers. 

\subsection{Heuristic Information Extraction}
The 5G Wireless Communication Protocol comprises both structured data (such as tables) and unstructured data (such as text). For structured data in the PDF version of the protocol, we employ Tabula \cite{mendes2017tabula} to extract tables embedded within the file, storing the extracted tables in a list of dataframes. In protocols like 5G, tables are integral data repositories, housing critical operational and management information, and encapsulating domain-specific terminologies. These terminologies, encapsulating complex protocol features and representing the unique language of this field, are structured within tables, presenting details on operational parameters, network configurations, routing information, device details, resource management, and security measures. To leverage the rich terminology within tables and maximize the use of information from the extracted dataframes, we collect all the words in the dataframes and apply a hierarchical filter to exclude unwanted words. The first filter is a general one, using stopwords to filter out generic words. The second is a domain filter, utilizing a term list provided by domain experts to filter out unrelated domain terms.

For unstructured data, we leverage Stanford CoreNLP \cite{manning2014stanford} to extract sources, targets, and relations. Our primary focus is on identifying identifiers and their formal properties, where the source and target represent the identifiers, and the relation symbolizes the formal properties between these identifiers. We utilize CoreNLP's Part-of-Speech (POS) tagging and dependency parsing pipelines in this process. POS tagging assigns suitable part-of-speech tags (such as noun, verb, adjective, etc.) to each token in a given text. Concurrently, dependency parsing evaluates the grammatical structure of a sentence based on the dependencies between words.

\begin{figure}[h]
\includegraphics[width=3.6in]
{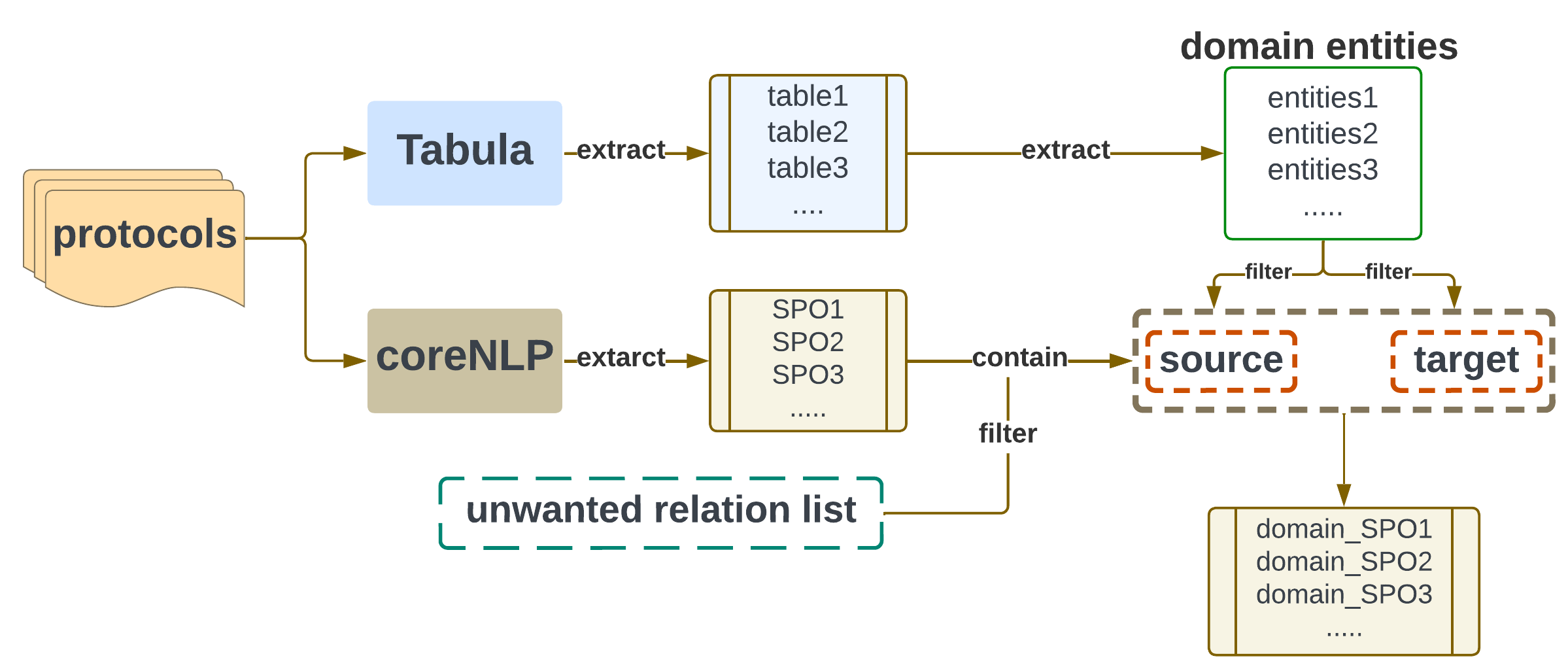}
\caption{Information Extraction Pipeline}
\label{tab:IE}
\vspace{-4mm}
\end{figure}

We apply the filtered domain entity list to the extracted source-relation-target (SRT) triples from CoreNLP, filtering the source and target tokens in the SRT triple with our entities of interest. Following the same method, we filter out unwanted predicates to specify the relations between identifiers as shown in Figure \ref{tab:IE}. Terminology extraction through Tabula and SRT extraction through coreNLP is illustrated in Information Extraction in Figure. \ref{fig:systemidagram} and detailed depicted in Figure.\ref{tab:IE}
\vspace{-3mm}

\subsection{Identifier and Formal Property}
Because our study is a pivot and preliminary research to analyze formal protocol utilizing natural language techniques, the main objective of this research is to propose an analysis framework and make a proof concept of the whole pipeline.  Consequently, we do not strictly define the identifiers used in this study. Instead, we adopt domain-specific terminologies from cybersecurity as our identifiers.

In the context of formal verification, for four main formal properties `Authentication', `Accounting',`Integrity',`Confidentiality' \cite{yang2023formal}, they may take the shape of state transition diagrams, preconditions, postconditions, invariants, and more abstract concepts used to mathematically model the protocol's operation. In communication protocols, these formats are written in natural language and may be classified into formal properties. For instance, in the sentence `The RRCReconfiguration includes the masterCellGroup containing the reportUplinkTxDirectCurrentTwoCarrier', the identifiers are `RRCReconfiguration' and `masterCellGroup'. The formal property connecting these two identifiers is `Integrity', but the word `includes' in the sentence represents `Integrity'. In this study, we do not train the model to classify formal properties directly. Instead, we predict the specific words representing formal properties. The representative words, provided by domain experts, are shown in Table \ref{tab:fp_label}.

\subsection{Hybrid Joint Sequence Labeling and Text Classification Task}
The main goal of our project is to extract the identifiers and their relations from the sentences in the communication protocol. The traditional NLP approach is to locate and extract the annotated tokens from sentences which is purely a sequence labeling task. Yet, we can construct the annotated dataset from communication protocol by using specific NLP tools such as coreNLP, the annotated dataset includes the identifiers and relations and their corresponding position information in the sentences. Considering the project goal of formal analysis based on communication protocol, we want to extract the identifiers and we want to know the relation or the formal property between the identifiers. So we convert the pure sequence labeling task into two sub-tasks: token classification task for identifiers extraction, and text classification for relation or formal proprieties prediction. 
For the token classification task of identifiers extraction, we follow the finetune-on-pre-trained models' pipeline. Pretrained models like BERT, RoBERTa, and GPT have been trained on massive amounts of text data. These models learn general features from the language like syntax, semantics, and word relationships. By fine-tuning these models on a specific task, we can leverage these previously learned features, allowing the model to perform well even with a smaller amount of task-specific data. Specifically, we used a pre-trained language model (PLM) and build a multi-layer perceptron (MLP) classifier on top of the PLM. 
For the text classification task of formal property prediction, we consider the information dependency of formal properties that depend on the identifiers also the context information for here which is the sentence. We concatenate the hidden state of the sentence and identifiers and pass the integrated latent representation to the formal property classifiers to predict the relation between the two identifiers under the context(sentence).

We designed three models to investigate the dependency of formal property on the identifiers and context. 
The first one is a disjoint training model in that we build two classifiers to separately extract identifiers and formal properties. The second one is jointly training the identifiers and formal properties, so we only consider the dependency between identifiers and formal properties. The third one is based on our proposed joint method which considers the dependency of formal properties. 


\begin{figure*}
     \centering
     \begin{subfigure}[b]{0.2\textwidth}       
        \includegraphics[width=\textwidth]{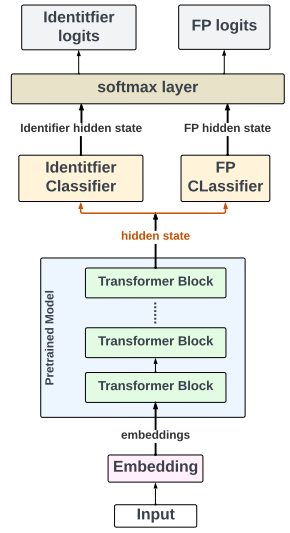}
        \caption{Disjoint Model}
        \label{fig:disjointflow}
     \end{subfigure}   
    \begin{subfigure}[b]{0.255\textwidth}
        \includegraphics[width=\textwidth]{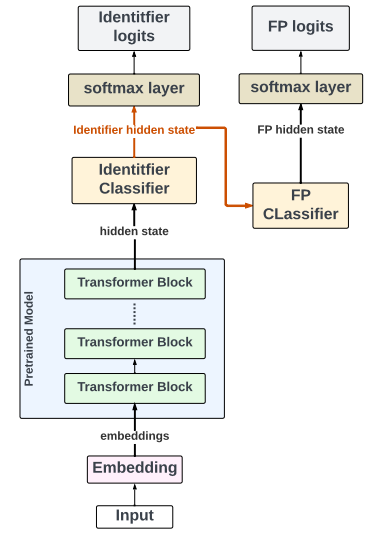}
        \caption{Joint1 Model}
        \label{fig:jointflow}
    \end{subfigure}  
    \begin{subfigure}[b]{0.25\textwidth}
        \includegraphics[width=\textwidth]{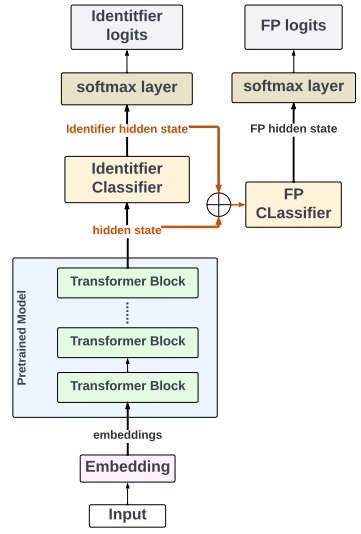}
        \caption{Joint2 Model}
        \label{fig:jointflow}   
    \end{subfigure}  
    \caption{Model Architectures}
    \label{fig:model_arch}
    \vspace{-15pt}
\end{figure*}

\subsubsection{Dataset}
In this study, we use the extracted and filtered SRT triples from core NLP and corresponding sentences as our annotation dataset. Our goal is to extract the identifiers and their corresponding formal properties. Therefore, we designate the source as 'identifier0', the target as 'identifier1', and the relation as the 'formal property'. Consequently, we have two entity types: 'identifier0' and 'identifier1'. As for relation types, we adhere to the list of interested relations provided by domain experts as shown in Table. \ref{tab:fp_label}. In our case, we have a total of 23 relations, thus yielding 23 relation types.
Given the influence of sequence length on NER task performance highlighted in prior research \cite{yuan2023information}, we prepend the corresponding sentence to the identifier triples, as shorter sequences have been shown to yield better results in named entity extraction.
\subsubsection{Model Selection}
In this study, we used transformer-based \cite{vaswani2017attention} Masked Language Model (MLM), since the bidirectionality property enables such a model to use context from both left and right to predict the masked word, which is crucial for accurately predicting each token's label in sequence labeling tasks. Also, we need the sentence information as another dependency factor of formal properties, the bidirectional understanding of MLM can provide a more holistic view of a sentence, capturing relationships and dependencies between all parts of the input and providing better sentence information. Combined the two factors, we used BERT as our base model, BERT \cite{devlin2018bert} is a pre-trained MLM, that has shown state-of-the-art performance on a wide array of NLP tasks, including sequence labeling tasks like NER \cite{liu2023pre} and text classification task which is beneficial for identifiers extraction and formal property classification in our project.

\subsubsection{Training Strategy} 
Our study aims to extract the identifiers and the formal property between the extracted identifiers under a specific context. Given their interdependent nature, to examine the inter-dependency of identifiers and formal properties, we propose three training strategies.

The first approach involves training classifiers for identifiers and formal properties separately. Specifically, we pass the hidden state of the sentence separately to the identifier classifier and the formal property classifier. We then train both classifiers in parallel to obtain their respective logits. This process is illustrated in Figure. \ref{fig:disjointflow}.
The second one is jointly training the identifiers and formal properties, that we first pass the identifiers to the classifier and use the logits of identifiers as input and pass the logits into formal property classifiers to get the formal property logits. This method did not consider the context information influence on the formal properties. 
The third one is based on our proposed method that builds the hybrid joint training model to extract the identifiers first. Then use the identifiers logits and sentence hidden states to predict the formal properties of the identifiers under the context (sentences). The algorithm is explained in Algorithm \ref{alg:joint} and model architecture is shown in \ref{fig:jointflow}. 
Since formal properties depend on the identifiers, even if two identifiers concurrently appear, under different contexts (sentences), the formal properties between the identifiers may change. Our hypothesis is that the third model that not only considers identifiers information but also takes context information to the formal properties prediction may provide the most representative information among the three models. Representative in this research is the prediction accuracy of formal properties.

\begin{algorithm}
      \textbf{Input} Sequence of tokens $X = \{x_1, x_2, ..., x_n\}$\\
     \textbf{Output} 
         Identifiers label $I = \{0,0,1,1,...\}$,\\ 
         Formal property label $P = \{0,0,0,10,...\}$
    \begin{algorithmic}[1]
        \Procedure{Joint\_Protocol\_Extraction}{$X$}
            \State  $E = embedding(X)$ \Comment{Embedding from pretrained model: BERT}
            \State  $H = pretrained\_model(E)$ \Comment{Get hidden state $H$ from pretrained model like BERT}
            \State  $I^{\prime} = identitfier\_classifer(H)$ \Comment{Predict the identifiers label from hidden state}
            \State  $I = argmax(softmax(I^{\prime}))$
            \State  $H^{\prime} = concatenate(H,I^{\prime})$
            \State  $P^{\prime} = fp\_classifer(H^{\prime})$   \Comment{Predict the formal property label from hidden state}
            \State  $P = argmax(softmax(P^{\prime}))$\\
            \Return $I = \{0,0,1,1,...\}, P = \{0,0,0,10,...\}$
        \EndProcedure
    \end{algorithmic}
\caption{Algorithm of Joint Identifier and Formal\_Property Extraction}
\label{alg:joint}
\end{algorithm}
\vspace{-4mm}

\section{Experiment Setup}
\subsection{Model Configuration}
We utilize 'bert-base-cased' as our selected BERT model, with the number of layers (L) set to 12, and a hidden dimension size of 768. Given our goal to leverage pre-trained knowledge and transfer it to our domain-specific task of in-domain token classification, we do not freeze the pre-trained layer. On top of the PLM, we add a classification layer with two MLPs for identifiers and formal properties, as depicted in Figures. \ref{fig:disjointflow} and \ref{fig:jointflow}.
The model configuration and hyper-parameter settings are shown in Table \ref{table:model_config}. 

\begin{table}[htbp]
\caption{Model Config and Hyperparameters Setting}
\centering
\catcode`,=\active
\def,{\char`,\allowbreak}
\renewcommand\arraystretch{1.2}
\begin{tabular}{p{3.4cm}<{\raggedright} | p{3.5cm}<{\raggedright} }
  \toprule
 \textbf{Model Config }         & \textbf{Hyperparameters}    \\ 
 \hline
  \midrule
    N\underline {o} of layers & 12  \\
    \hline
    Hidden Dimension      & 768     \\
    \hline
    Learning rate     & 1e-4        \\
    \hline
    num\_epochs       & 20 (40 only for Joint1)        \\
    \hline
    Optimizer         & AdamW       \\ 
    \hline
    weight\_decay     & 0.1       \\
    \hline
    no\_regularization layer  & "bias", "ln\_1", 'ln\_2' \\
    \hline
  \bottomrule
\end{tabular} 
\label{table:model_config} 
\end{table}


\subsection{Raw Input Documents to Dataset Construction}
The source document that we used to analyze in this research is `5G NR Radio Resource Control (RRC) Protocol specification' \cite{3gpp.38.331}. This document outlines the procedures and command formats associated with 5G communication at the RRC layer, serving as a crucial source for understanding and exploring 5G communication protocols.

\subsubsection{Identifier Categorization}
For the filtered source and target elements in the SRT triples derived from CoreNLP (as shown in Figure \ref{tab:IE}), we label the source as `identifier0' and the target as `identifier1' in each sentence. Tokens not classified as identifiers are categorized as `O'. Therefore, we have two types of identifiers, and the total number of token labels is three, including the non-identifier tokens. 

\subsubsection{Formal Property Categorization}
We use the formal property list provided by domain experts for categorization, as shown in Table \ref{tab:fp_label}. There are 23 formal property words, meaning we have 24 types of labels, including non-formal property tokens represented by `O'.

\begin{table}
\caption{Predicate Convert Formal\_Property}
\centering
\begin{tabular}{p{0.7\linewidth} | p{0.2\linewidth} }
    \hline
    \textbf{Formal Property Word List} & \textbf{Formal Property Category}\\
    \hline
    `decode',`encode' & Confidentiality \\
    \hline
    `verify' &  Integrity\\
    \hline
    `access', `reestablish' & Authentication\\
    \hline
    `count'&  Accounting\\
    \hline
    `build',`complete', `append',`belong',`store',`contain', 
    `include',`combine' & Belong\\
    \hline
    `imply',`establish',`modify',  
    `denote',`utilize',`set',
    `change',`define',`=' & Generation\\
    \hline
\end{tabular}
\label{tab:fp_label}
\vspace{-6mm}
\end{table}

\subsection{Training and Evaluation}

We employ a hierarchical regularization strategy to avoid over-penalizing during training, which might lead to an under-fitting issue. For all bias modules, MLP layers, and linear layers in the multihead\_attnetion component of the transformer block, we refrain from regularization during training. The details are elaborated in Table \ref{table:model_config}. All results are presented in the ensuing results section.

We adhere to the standard evaluation metrics used in sequence labeling tasks, which involve computing the ratio of correctly extracted identifiers and non-identifier tokens to the total number of tokens in the sequence: $acc\_idf = (n_{(correct\_idf)}\quad+\quad\ n_{(incorrect\_idf)})/n_{tokens}$. This measurement provides an overview of the model's performance in terms of correctly identifying both the entities (identifiers in our case) and other tokens.  The $n_{(correct\_idf)}$ denotes the total number of correctly extracted identifiers in a batch, $n_{(incorrect\_idf)}$ denotes the total number of correctly extracted non-identifiers in a batch, $n_{tokens}$ is the total number of tokens in a batch. 

For formal properties, we adopt a slightly different evaluation metric. Given the complexity and potential multi-label nature of formal properties, we evaluate the performance by $acc\_fprop = n_{(correct\_fprop)}/ batch\_size$, which is the ratio of correctly predicted properties to the total batch size, and the$n_{(correct\_fprop)}$ is the correct predicted formal properties in a batch, `batch\_size' is the number of samples in a batch. 


\section{Result Analysis}

\noindent\textbf{Heuristic Information Extraction:} In the information extraction task, we specifically concentrated on two sub-tasks: extracting terminologies and performing SRT extraction. The extracted terminologies are used as a dictionary to further filter the SRT triples. The filtered SRT triple will only contain the in-domain terminologies in the source and target. 

\noindent\textbf{Terminology Extraction: } For this phase, we used a 5G communication protocol spanning $1197$ pages as our source document. Utilizing tabula, we were successful in extracting $696$ tables from this document. These tables were then serialized into lists of words, which subsequently allowed us to extract a total of $10905$ terminologies.
\subsubsection{Identifier and Formal Property Extraction}
By using coreNLP, we extract $19586$ source-relation-target-sentence groups from the documents. We incorporated the terminologies derived from the previous step, adhered to a hierarchical filtering strategy, and applied lemmatization. This process yielded $16244$ groups.

The formal property list provided by the domain expert is as in table \ref{tab:fp_label}, so the total formal\_property type is 24 including 'O' which stands for not formal\_property. By filtering for the formal\_property type of relation within the original SRT groups, we retained $2495$ SRT groups. Since the majority of the sentences within these groups contained fewer than 150 tokens, we removed sentences exceeding 200 tokens from the groups and set a maximum sequence length of $256$ in BERT. Consequently, we ended up with $1486$ source-relation-target-sentence groups. 

\begin{figure}
     \centering
     \begin{subfigure}[b]{0.24\textwidth}
        \includegraphics[width=\textwidth]{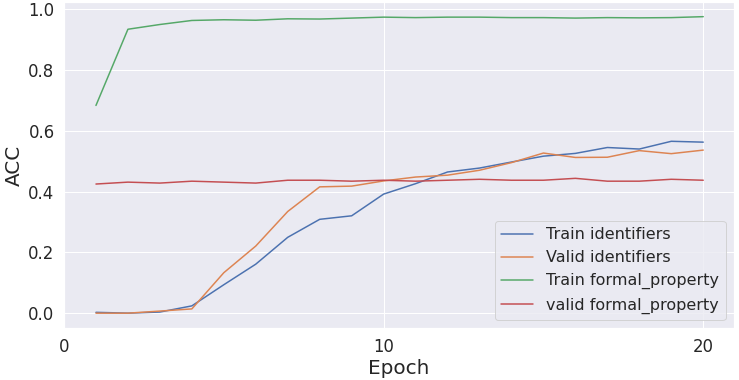}
        \caption{Disjoint Training Acc}
        \label{fig:Acc_app1}
    \end{subfigure}   
    \hfill
    \begin{subfigure}[b]{0.24\textwidth}
        \includegraphics[width=\textwidth]{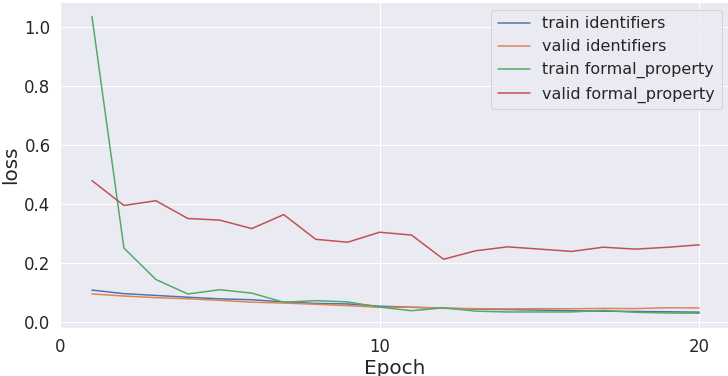}
        \caption{Disjoint Training Loss}
        \label{fig:Loss_app1}     
    \end{subfigure}
    \caption{Disjoint Training Accuracy}
    \label{fig:disjoint_result}
    \vspace{-2mm}
\end{figure}

\noindent\textbf{Hybrid Training Disjoint Result: : }In Disjoint training, we separately train the identifier classifier and the formal property classifier in parallel, which means the loss in formal properties did not inherit logits from identifiers. From the loss result, we see that the identifiers' loss converged, but the formal property loss did not converge as showed in Figure.\ref{fig:Loss_app1}. For accuracy, identifiers extraction accuracy is almost 60\% in Figure. \ref{fig:Acc_app1}. We have 3 labels for identifiers, the random guess is 30\% for each label. But for formal properties prediction, the result is in line with the loss which means overfit for FP training which indicates an overfitting problem.

\begin{figure*}
     \centering
     \begin{subfigure}[b]{0.32\textwidth}      
        \includegraphics[width=\textwidth]{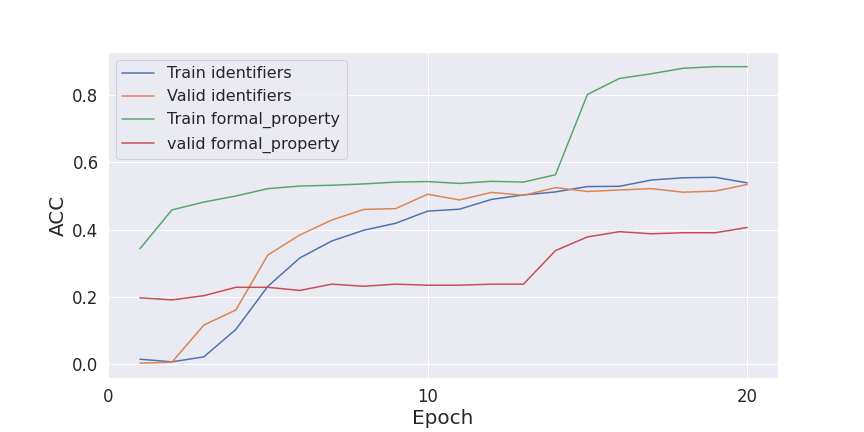}
        \caption{Joint1 20 Epoch Training Accuracy}
        \label{fig:Acc_app2}
     \end{subfigure}    
    \begin{subfigure}[b]{0.276\textwidth}
        \includegraphics[width=\textwidth]{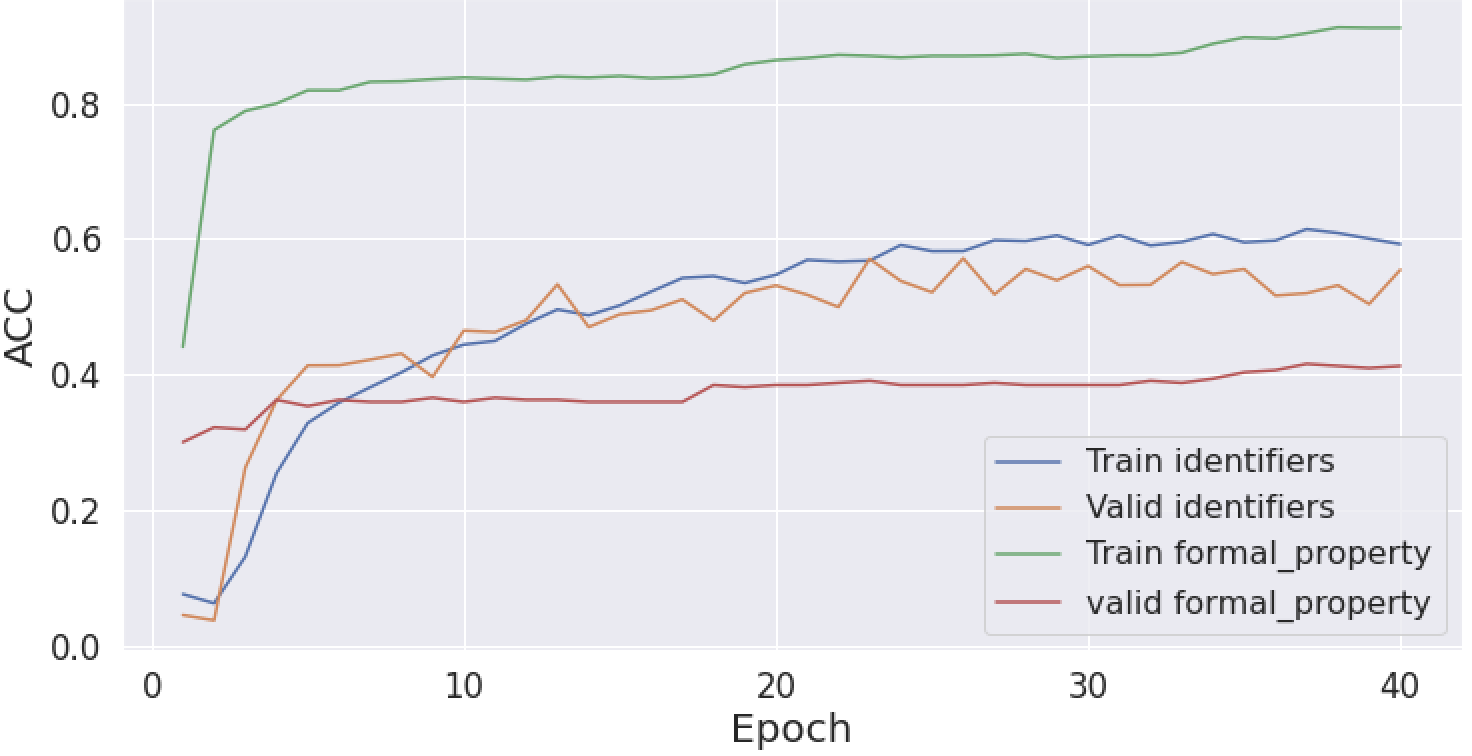}
        \caption{Joint1 40 Epoch Training Accuracy}
        \label{fig:Acc_app2_40}
     \end{subfigure}
    \begin{subfigure}[b]{0.32\textwidth}
        \includegraphics[width=\textwidth]{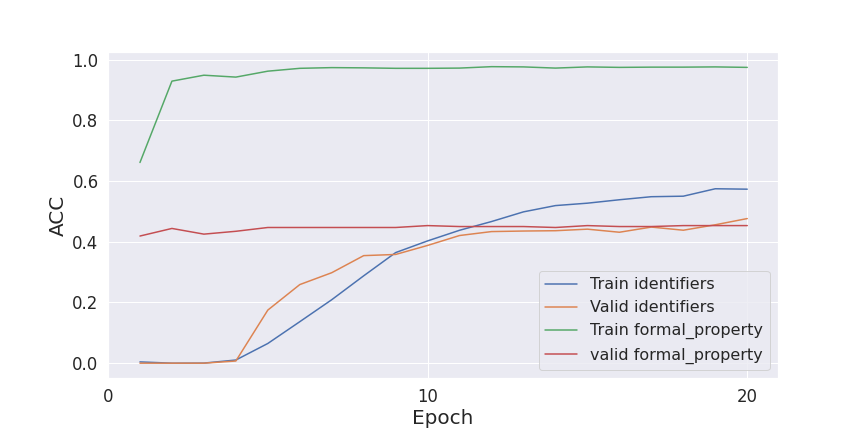}
        \caption{Joint2 20 Epoch Training Loss}
        \label{fig:Acc_app3}     
    \end{subfigure}
    \caption{Joint Training Result}
    \label{fig:joint_result}
    \vspace{-15pt}
\end{figure*}

\noindent\textbf{Hybrid Training Joint Result: }
In the joint training process, we implemented two joint strategies to investigate the sentence influence on formal property prediction. The first model (Joint1) did not take the sentence hidden state to predict formal property, we only pass the identifier logits after the identifier classifier to the formal property classifier to get the formal property logits and make the final formal property prediction. In the second joint model (Joint2), we concatenate the sentence hidden state and identifier logits and pass the merged hidden information to the formal property classifier. On comparing, we found the formal property prediction accuracy is much lower in Joint1 than in Joint2, but after around 12 epochs, we found the formal property prediction in Joint1 reached a similar performance as Joint2 which indicates the sentence influence may accumulate during training and make contribute to the formal property prediction through the identifiers' logits. Another thing that needs to be noticed is in Joint1 the accuracy is still in an increasing trend, but in Joint2, the performance for both identifiers and formal properties seems static. So we extend training sessions from 20 to 40 of the Joint1 strategy, and the result is shown in Figure. \ref{fig:Acc_app2_40}. Results show that both the loss and accuracy improved. But the prediction performance of formal properties is with limitation. Partially because the formal properties training depends on the identifiers training, which still faces the overfitting problem. So the limited improved identifier training also restricts the formal property training.

\begin{table}[htbp]
\caption{Identifier \& Formal Property Training Accuracy}
\vspace{-1mm}
\centering
\begin{tabular}{|p{1.4cm}||p{0.9cm}|p{1.5cm}|p{0.9cm}|p{1.5cm}|}
    \hline
    Train\_Strategy 	&Train\_Idf &Train\_Property &Valid\_Idf &Valid\_Property\\

    \hline
    Disjoint	  &0.3339	&0.9536	&0.3993	&0.4351	\\
    Joint1	      &0.3773	&0.6170	&0.5343	&0.2799 \\
    jpint1\_40	  &0.4029	&0.9050	&0.3927	&0.4253	\\
    Joint2        &0.3356	&0.9523	&0.3127	&0.4459	\\
    \hline
\end{tabular}\\
\label{tab:train_results}
\vspace{-3mm}
\end{table}

\begin{figure}
     \centering
     \begin{subfigure}[b]{0.35\textwidth}       
        \includegraphics[width=\textwidth]{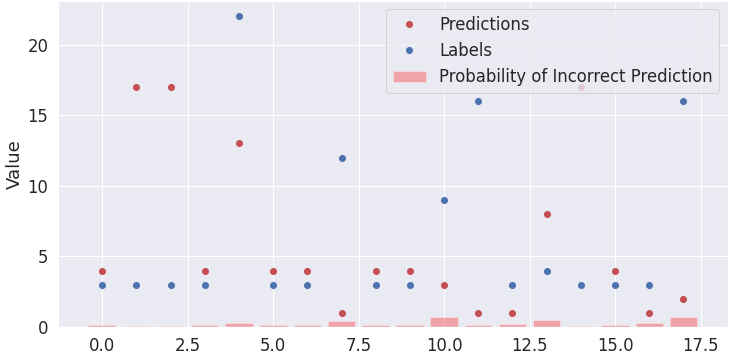}
         \vspace{-3mm}
        \caption{Joint1 Incorrect Formal Property Prediction}
        \label{fig:app2_incorrect_fp}
     \end{subfigure}    
     
    \begin{subfigure}[b]{0.35\textwidth}
        \includegraphics[width=\textwidth]{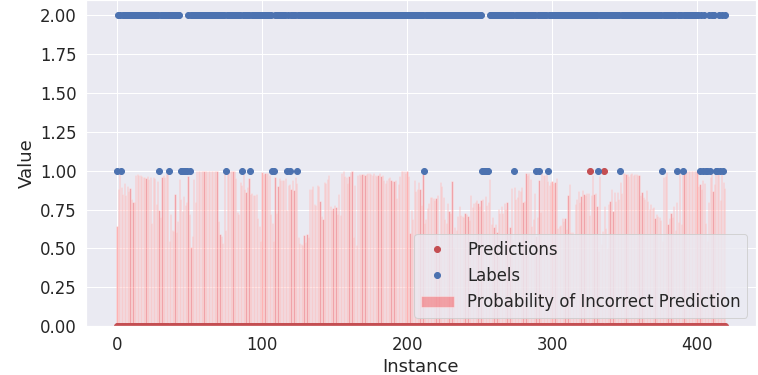}
        \vspace{-5mm}

        \caption{Joint1 Incorrect Identifier Prediction}
            \vspace{-2mm}

        \label{fig:app2_incorrect_idf}
    \end{subfigure}

    \caption{Joint1 Incorrect Prediction}
        \vspace{-8mm}
\label{fig:joint1_result_40}
\end{figure}

Further, we investigate the incorrect prediction distribution of the Joint1 model. We extract the incorrect perdition result from the last training epoch. We noticed that for the identifier incorrect prediction, most of the incorrect prediction is non-identifiers in which the ground truth label should be `1' or `2', but incorrectly predicted to non-identifier label `O' as shown in Figure. \ref{fig:app2_incorrect_idf}. This result may indicate an unbalanced label in the identifiers because most of the token labels in sentences are labeled `O', the non-identifier,this may cause very unbalanced label distribution in the training. In formal property incorrect predictions, we found that most of the wrongly predicted labels are `3' which predicted to `4' or `17' as shown in Figure. \ref{fig:app2_incorrect_fp}. Based on partial formal property distribution in Table. \ref{tab:fp_stat}, we noticed the incorrect prediction of formal property is not due to unbalanced data, this may result from the identifier dependency, which if the identifier prediction is wrong, the formal property will also predict incorrectly. We will discuss this in the following section. 

\begin{table}
\vspace{-2mm}
\caption{Formal\_property\_Stats (partial)}
\centering
\begin{tabular}{p{0.22\linewidth} | p{0.31\linewidth} |p{0.35\linewidth}}
    \hline
    \textbf{formal property} & \textbf{formal property IDs} & \textbf{formal property Counts}\\
    \hline
    include & 3	 & 620 \\
    \hline
    contain	& 4 &  327\\
    \hline
    utilize & 17 & 10\\
    \hline
\end{tabular}
\label{tab:fp_stat}
\vspace{-4mm}
\end{table}


\section{Discussion and Limitation}

\subsection{Heuristic Terminology and Semantic Roles Extraction}
In the preparation of the dataset, we employed Stanford's CoreNLP to extract SRT triples, utilizing its deterministic PTBTokenizer for tokenization. For the construction of the model, the BERT architecture was selected as the framework. Both require the processing of text into individual tokens. However, they differ significantly in their tokenization methodologies. While CoreNLP employs PTBTokenizer, which operates on a deterministic basis, BERT makes use of a word-piece tokenizer that subdivides words into subword units. This discrepancy in tokenization between PTB and the word piece can result in differing token representations for the same lexical item, adding a layer of complexity to the alignment and integration of these two components within our research workflow. 

\subsection{Training Result and Analysis}
The results from both the disjoint and joint training approaches revealed that the joint training strategy outperforms the disjoint method in both an identifier and formal property tasks. The reason lies in the sequential architecture of the joint training model. In this model, the gradient from the final loss first passes through the formal property classifiers and then both the identifier classifier and the pre-trained BERT layers during back-propagation. In contrast, even though the identifier and formal properties classifiers share the same final loss in disjoint training, the gradient passes equally to both classifiers. The formal properties loss enhances the final loss that generates the gradient, but in disjoint training, the identifier and formal property classifiers don't share sequential back-propagation. Consequently, the identifier classifier learns more through the formal properties loss, which influences the identifier classifier's learning behavior. 
For our extended training of the Joint1 model, we noticed some improvement compare to the Disjoint model and Joint2 model. But the improvement is not significant, the main reason behind is an overfitting problem. We tried several training strategies to alleviate this problem, such as increasing the dropout rate, using higher regularization weight in the optimizer, increasing model complexity, etc, but did not see significant improvement. This indicates the overfitting problem is not just due to model complexity or experiment setup, whereas dataset size may be the most crucial reason since we only have around $1400$ training samples. Based on the result, in the future, we will construct more data and implement the same training strategy to investigate the performance. Another aspect that may influence the overall performance is unrestricted identifiers and formal properties, this problem may be trivial since our goal is to provide a proof of concept of the pipeline. We plan to employ a more fine-grained dataset to refine this procedure in future studies. 

For the incorrect predictions, we found unbalanced label distribution present in the identifier and consequently influence formal properties predictions. In the future, we will mask out the non-identifier tokens during back-propagation which will provide more guided loss on the identifiers and may also improve the prediction of formal property tokens. 

\vspace{-2mm}
\subsection{Disjoint and Joint Training}
We used the averaged result to represent the performance of each model in Table \ref{tab:train_results}. The experimental results underscore the superior performance of the Disjoint model, which trains identifiers and formal properties in parallel. However, it is crucial to recognize the inherent complexity of our project. While both identifiers and formal properties are dependent on the sentence structure, the latter also relies heavily on the former. Neglecting the influence of identifiers on formal properties could lead to significant design flaws in the experiment. Such oversight may create a deceptive impression of improved training performance, masking underlying issues that could undermine the integrity of the model. In future research, we will delve into the model architecture and find the optimal solution to balance sentence and identifier influence on formal properties' prediction. 

\section{Conclusion and Future Research}
In this paper, we introduce an innovative approach to performing formal verification analysis using NLP-assissed techniques. Our method involves an automated process pipeline for extracting terminologies and semantic roles from domain-specific documents, which subsequently form the foundation of our annotation training dataset. Moreover, we employ a transformer-based language model as the backbone for our classifiers, further enhancing the efficacy of our approach. Upon comparing the results from disjoint and joint training strategies, we find that only passing the identifier logits to extract formal properties slightly outperforms other methods. This proposed pipeline illustrates the potential of our novel approach in the realm of formal verification analysis.

This study serves as a fundamental exploration of the application of NLP for formal verification analysis, showcasing our preliminary experimental results. As we continue to develop our approach, we aim to fine-tune our methodologies, address identified challenges, and enhance the performance and versatility of our model in subsequent research pursuits. Our future efforts will focus on refining the dataset with ground-truth identifiers and verified formal properties. With regards to the model, we intend to persist with the joint approach, while seeking solutions to further improve the accuracy of identifier extraction and FP prediction, thus ensuring that our methodology remains robust and adaptable to diverse formal verification analysis tasks. 
\vspace{-2mm}
\section*{Acknowledgment}
This effort was sponsored by the Defense Advanced Research Project Agency (DARPA) under grant no. D22AP00144. The views and conclusions contained herein are those of the authors and should not be interpreted as necessarily representing the official policies or endorsements, either expressed or implied, of DARPA or the U.S. Government.

\bibliographystyle{IEEEtran}
\bibliography{myref}




\end{document}